\documentclass[sigconf, screen]{acmart}

\usepackage{kotex} 
\newcommand{\figref}[1]{Fig.~\ref{#1}}

\newcommand{\tabref}[1]{Table~\ref{#1}}

\newcommand{\secref}[1]{Sec.~\ref{#1}}
\newcommand{\Figref}[1]{Figure~\ref{#1}}

\newcommand{\ie}{\textit{i}.\textit{e}.~}
\newcommand{\eg}{\textit{e}.\textit{g}.~}

\newcommand{\ourmodel}{ReFu}
\usepackage{multirow}

\AtBeginDocument{%
  }

\setcopyright{acmlicensed}
\copyrightyear{2022}
\acmYear{2022}

\acmConference[MM '22]{Proceedings of the 30th ACM International Conference on Multimedia}{October 10--14, 2022}{Lisboa, Portugal}
\acmBooktitle{Proceedings of the 30th ACM International Conference on Multimedia (MM '22), October 10--14, 2022, Lisboa, Portugal}
\acmPrice{15.00}
\acmISBN{978-1-4503-9203-7/22/10}
\acmDOI{10.1145/3503161.3547971}

\begin{document}

\title{ReFu: Refine and Fuse the Unobserved View for Detail-Preserving Single-Image 3D Human Reconstruction}

\author{Gyumin Shim}
\authornote{Both authors contributed equally to this research.}
\author{Minsoo Lee}
\authornotemark[1]
\affiliation{%
  \institution{Korea Advanced Institute of Science and Technology}
  \city{Daejeon}
  \country{South Korea}
  }
\email{{shimgyumin, alstn2022}@kaist.ac.kr}


\author{Jaegul Choo}
\affiliation{%
  \institution{Korea Advanced Institute of Science and Technology}
  \city{Daejeon}
  \country{South Korea}}
\email{jchoo@kaist.ac.kr}








\renewcommand{\shortauthors}{Gyumin Shim, Minsoo Lee \& Jaegul Choo}

\begin{abstract}
Single-image 3D human reconstruction aims to reconstruct the 3D textured surface of the human body given a single image.
While implicit function-based methods recently achieved reasonable reconstruction performance, they still bear limitations showing degraded quality in both surface geometry and texture from an unobserved view.
In response, to generate a realistic textured surface,
we propose \ourmodel, a coarse-to-fine approach that {\it refines} the projected backside view image and {\it fuses} the refined image to predict the final human body.
%
%
To suppress the diffused occupancy that causes noise in projection images and reconstructed meshes, we propose to train occupancy probability by simultaneously utilizing 2D and 3D supervisions with occupancy-based volume rendering.
%
We also introduce a refinement architecture that generates detail-preserving backside-view images with front-to-back warping.
Extensive experiments demonstrate that our method achieves state-of-the-art performance in 3D human reconstruction from a single image, showing enhanced geometry and texture quality from an unobserved view.
\end{abstract}


\begin{teaserfigure}
\begin{center}
  \includegraphics[width=0.9\linewidth]{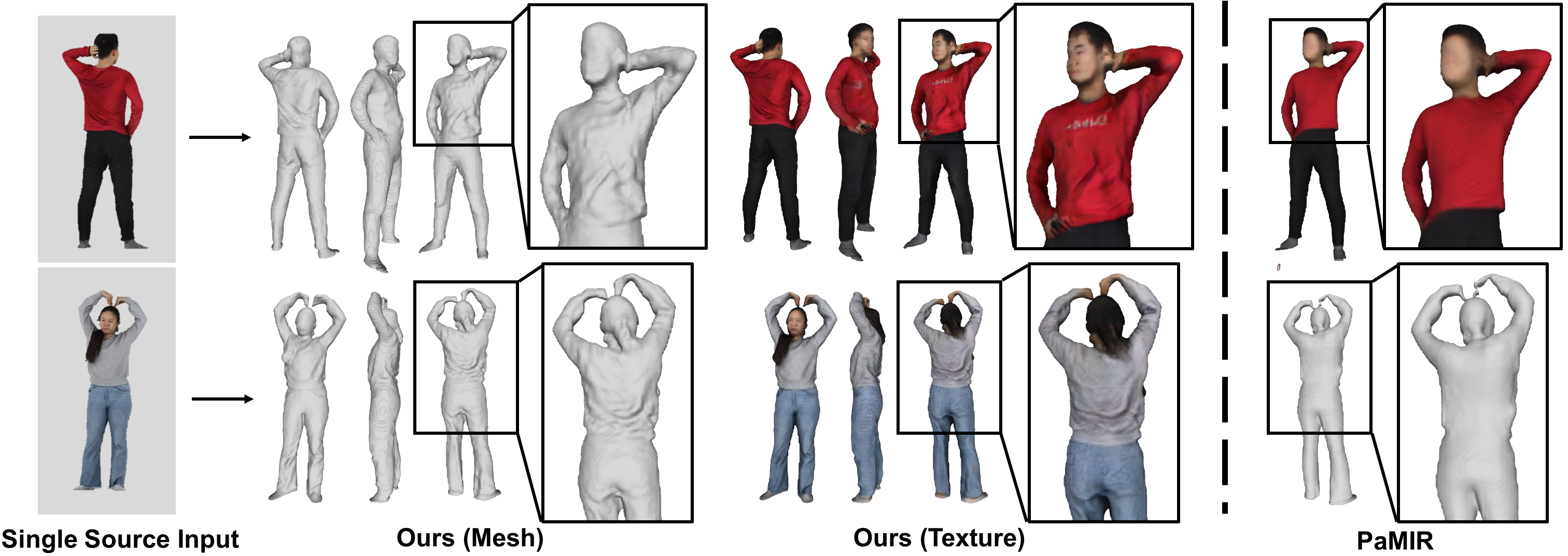}
  \caption{Qualitative results of our method in terms of surface geometry and texture. Given the source image (first column), our method results in realistic textured meshes with high-frequency details. In contrast, PaMIR shows over-smoothed outputs from an unobserved view.}
  \label{fig:teaser}
\end{center}
\end{teaserfigure}

\maketitle





\section{Introduction}
\label{sec:intro}
3D human reconstruction has received substantial attention due to its importance and practicality in VR/AR content and metaverse domains. 
Image-based 3D human reconstruction aims to predict the textured human mesh given a single or multiple images. 
However, recovering a realistic 3D human from a single image is an ill-posed problem due to depth ambiguity and occlusions.
Also, the high variance of human pose and clothes worsens the problem.

Recently, numerous deep-learning-based methods~\cite{natsume2019siclope, saito2019pifu, zheng2021pamir, zheng2019deephuman, varol2018bodynet, alldieck2019tex2shape, jackson20183d, pavlakos2019expressive, zhu2019detailed} have been proposed to reconstruct a 3D human body from a single image. 
In the early stages, leveraging the pre-defined human template mesh, \eg SMPL~\cite{loper2015smpl}, several methods~\cite{pavlakos2019expressive, alldieck2019tex2shape, zhu2019detailed} learn to deform the template mesh conditioned on a given source image. 
Recently, among non-parametric representations such as voxels\cite{varol2018bodynet, jackson20183d}, silhouettes\cite{natsume2019siclope}, and depth maps~\cite{gabeur2019moulding}, methods based on implicit representations have achieved a paradigm shift in 3D human reconstruction tasks, showing substantial improvement for representing the textured surface of clothed humans. 
Implicit-function-based methods~\cite{saito2019pifu, saito2020pifuhd} conditioned on the pixel-aligned feature of a source image learn to estimate occupancy fields and corresponding textures.
PaMIR~\cite{zheng2021pamir} further improves reconstruction accuracy by combining the volumetric feature from SMPL with condition variables.

\begin{figure}[t]
\captionsetup{width=1\linewidth}
\begin{center}
\begin{tabular}{@{}c}
\includegraphics[width=1.0\linewidth]{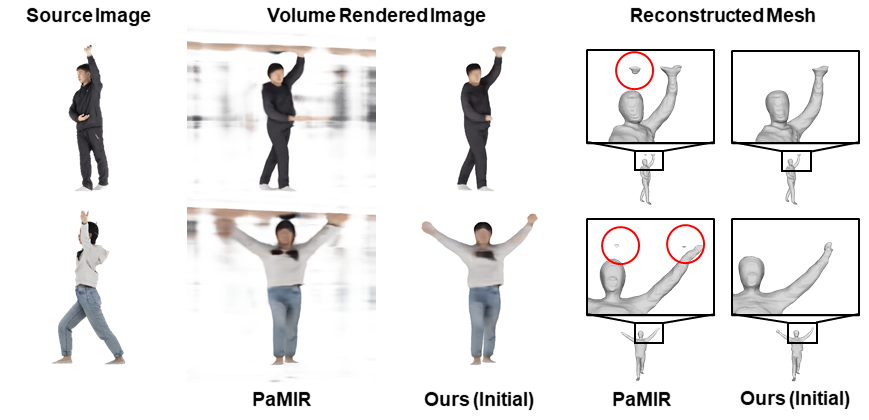} 
\end{tabular}
\end{center}
\caption{\small Volume-rendered images and reconstructed meshes of PaMIR~\cite{zheng2021pamir} and ours. 
While PaMIR presents artifacts in the volume-rendered images and the meshes, our initial stage shows the clear projection images and the reconstructed meshes.
}
\label{fig:volmuerender}
\end{figure}

However, from a view not observed in the given source image, the quality of the surface geometry and the texture are significantly degraded with an over-smoothed mesh surface and a lack of high-frequency texture details. 
We argue that this limitation is due to the per-point loss function, \eg the L1 Loss and the mean squared error which encourage the point-wise averages of plausible solutions causing over-smoothed results~\cite{isola2017image, srgan}.
To generate a realistic textured mesh from any view, we propose a staged method where we predict a coarse human body ({\it initial stage}), refine the coarse image projected from the unobserved view ({\it refinement stage}), and fuse the refined image with the input condition to reconstruct the final human body ({\it fusion stage}).
To obtain a projected coarse image from a 3D implicit function, we adopt a volume rendering process, \ie ray-tracing techniques for compositing colored density, that is proven to be effective in synthesizing high-quality images. 
However, as shown in \figref{fig:volmuerender}, the volume-rendered images of PaMIR~\cite{zheng2021pamir}, the state-of-the-art 3D human reconstruction method, are noisy and blurry.
This undesirable noise stems from the diffused occupancy, \ie non-zero occupancy value outside the object surface, that blocks the object surface.
Also, the diffused occupancy can cause floating artifacts in the reconstructed meshes.
To suppress the diffused occupancy around the object, we propose to train the occupancy probability by simultaneously utilizing 2D and 3D supervisions with the occupancy-based volume rendering.
To achieve this, we interpret alpha values in the cumulative rendering process~\cite{mildenhall2020nerf} as occupancy probabilities that are trained with 3D supervisions.
Since the higher transmittance value is computed at the diffused occupancy area, the volume rendering loss is efficient in lowering the entropy of the occupancy values.
Thanks to the volume-rendering-combined 3D supervised learning, we can acquire noiseless occupancy fields and a clear projected image from any target view.

Leveraging a clear projection image, we introduce architecture that generates a realistic image from an unobserved view, \ie backside, in a coarse-to-fine manner.
To refine the coarse image with high-frequency details, we transfer the source local feature to the target image using the front-to-back warping function.
The function can be obtained by volume rendering the 3D query points with the value of the source grid coordinates. 
%
%
The rationale behind this warping function is that the corresponding coordinates between the front and backside view share semantically similar visual features such as the texture of clothes and skin color.
By transferring the local feature of the front view to the backside and injecting it with spatial modulation~\cite{albahar2021pose, park2019semantic}, we extend the StyleGAN generator~\cite{karras2020analyzing} that takes a coarse image as input and generates a realistically refined image.
Using the refined backside-view images as an additional input, our final model can predict a perceptually enhanced 3D human mesh.
We demonstrate that the quality of the mesh and texture outperforms the previous state-of-the-art methods.


In summary, our contributions are as follows:
\begin{itemize}
\setlength\itemsep{0em}
    \item We propose to train occupancy probabilities by leveraging both 2D and 3D supervisions with the occupancy-based volume rendering.
    \item We also propose the refinement architecture that can generate high-frequency details in the backside-view image by warping local features from the source to the target view.
    \item We demonstrate that the quality of single-image 3D human reconstruction is improved for both geometry and texture by fusing the refined backside-view image.
\end{itemize}

\section{Related Work}

\subsection{3D Human Reconstruction}

Previous studies focus on reconstructing 3D human meshes using video sequences~\cite{alldieck2018detailed, alldieck2018video} or multi-view RGB-D data~\cite{yu2018doublefusion, zheng2018hybridfusion}. 
While these approaches show reasonable 3D reconstruction qualities by addressing the depth ambiguity of shapes, they do not work well against only single or sparse input images.   
To reconstruct human meshes from a single image, several methods~\cite{loper2015smpl, pavlakos2019expressive} utilize parametric body models that provide a strong prior knowledge of the human body. 
To represent fine details of the surface of the parametric human body, other studies~\cite{alldieck2019tex2shape, lazova2019360, lahner2018deepwrinkles} attempted to estimate displacements in a pre-defined UV space. 
Methods~\cite{varol2018bodynet, jackson20183d} that combine a voxel representation with the parametric body model further improve the reconstruction quality. 
However, such parametric-model-based methods still show severe limitations on predicting surfaces that have large deformations such as loose clothes or hairs. 
Also, the high memory cost of voxel limits the resolution, which leads to the degradation of the fine details of mesh surfaces. 

Compared to explicit representations such as voxel grids~\cite{maturana2015voxnet}, point clouds~\cite{qi2017pointnet,lin2018learning}, and meshes~\cite{groueix1802atlasnet, wang2018pixel2mesh}, implicit representations~\cite{park2019deepsdf, mildenhall2020nerf} define the 3D surface with continuous probability fields.
As implicit representations have been widely utilized for 3D object representation due to their memory efficiency and continuity, they are also widely utilized in 3D human reconstruction~\cite{saito2019pifu, saito2020pifuhd, zheng2021pamir, he2020geo}. 
Recently, PIFu~\cite{saito2019pifu} introduced a pixel-aligned implicit function that represents the continuous occupancy and color value for each 3D query point conditioned on the corresponding 2D image feature. 
PIFu-HD~\cite{saito2020pifuhd} and Geo-PIFu~\cite{he2020geo} further improved the reconstruction quality level of detail by formulating coarse-to-fine approaches. 
PaMIR~\cite{zheng2021pamir} combines a parametric human body model with the implicit function to improve the performance under the severe occlusions or large pose variations.
Although these recent methods show a reasonable reconstruction quality of surface geometry and texture, there are still limitations in representing fine details from the backside view. 
To overcome the degraded quality from unobserved views, we propose \ourmodel~ that refines and fuses the backside view to enhance the final output result.

\subsection{Condition-guided Human Image Synthesis}
Condition-guided human image synthesis has attracted a substantial amount of attention due to its wide range of applications, \eg motion transfer, human reposing, and virtual try-on.
This task aims to translate a source human image to a target image in which the shape and appearance are changed according to given conditions, \eg viewpoint, input pose, and target clothes.
Inspired by spatial transformer networks~\cite{jaderberg2015spatial}, recent works predict a spatial transformation and transfer the source image pixel/feature to a target image space to preserve the high-frequency details.
The warping function is predicted by conditions such as skeleton pose~\cite{zhu2019progressive, ren2020deep}, UV parameterization pose~\cite{sarkar2021style, albahar2021pose, yoon2021pose}, and human parsing map~\cite{yu2019vtnfp, choi2021viton, xu20213d}.
A state-of-the-art human reposing method~\cite{albahar2021pose} predicts the warping function from the UV parameterization pose~\cite{guler2018densepose} and successfully renders a photo-realistic human image leveraging the StyleGAN~\cite{karras2019style} architecture with the warped source feature.

Although we adopt the condition-guided StyleGAN architecture for refinement, our method differs from the previous work~\cite{albahar2021pose} in two aspects.
First, our method is conditioned on the coarse backside-view image obtained in advance by projecting our 3D human implicit function that contains the pose information and the approximate texture.
Second, we compute the front-to-back warping function in a 3D-aware manner by projecting source grid coordinates to the backside view.
In this way, the coarse backside-view image is refined into the photo-realistic image that is utilized for generating the detail-preserving 3D human body.  

\begin{figure*}[t]
\centering
\begin{tabular}{@{}c}
\includegraphics[width=1\linewidth]{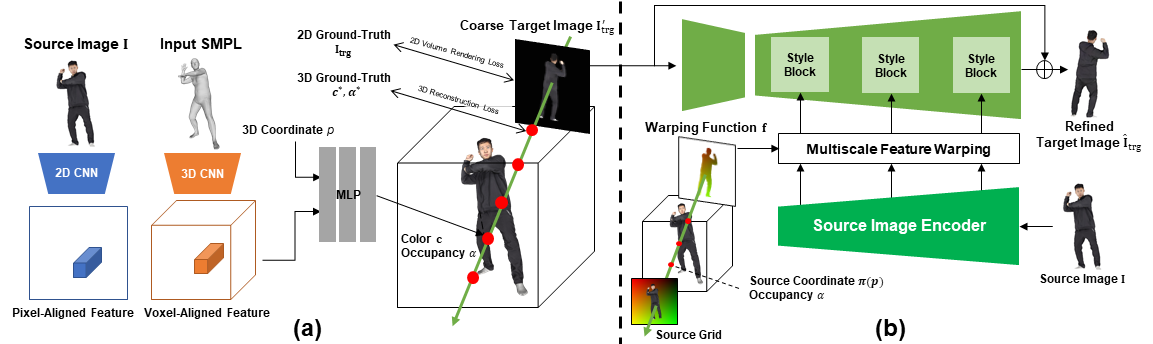} \\
\end{tabular}
\caption{\small Overview of our approach. 
(a) denotes the initial stage that predicts the coarse textured surface.
(b) indicates the refinement stage that takes the projected backside-view image as an input and generates a realistic image with warped source features.
Fusing the refined target image with the source image, our fusion network predicts the final human body using the identical architecture of the initial stage.
}
\label{fig:framework}
\end{figure*}

\section{Proposed Method}

An overview of our proposed method is illustrated in \Figref{fig:framework}. 
We propose a three-stage approach that consists of 1) the initial stage,
2) the refinement stage, and 3) the fusion stage. 
Given a single image {\bf I} as input, our initial stage predicts the coarse textured surface. 
In the next stage, we refine the image projected from the backside view where the initial geometry and texture lack reality.
Lastly, the fusion network reconstructs the final textured mesh by incorporating the refined backside-view image into the input conditions.

\subsection{Preliminary : SMPL}

Following previous studies~\cite{zheng2019deephuman, zheng2021pamir}, the parametric human body model, SMPL~\cite{loper2015smpl}, is utilized to extract volumetric features from an input image. 
SMPL is a function M($\cdot$) that maps pose $\theta$ and shape $\beta$ to a mesh of 6890 vertices. 
Given the template model ${\mathbf T}$, we obtain a posed shape instance as 
\begin{equation}
\begin{gathered}
M(\beta, \theta)= W(T(\beta, \theta), J(\beta), \theta, {\bf W}),\\
T(\beta, \theta)= {\bf T} + B_s(\beta) + B_p(\theta),
\end{gathered}
\end{equation}
where $W$ denotes a linear blend-skinning function with the weight $\bf W$, $B_p(\theta)$ and $B_s(\beta)$ are pose-dependent and shape-dependent deformations from the T-shape template model {\bf T}, respectively, and J($\cdot$) denotes the skeleton joints. As proven in previous work~\cite{zheng2019deephuman, zheng2021pamir}, the SMPL model fit to the input image provides a strong prior with respect to the pose and shape of a human body. 
Among recent studies~\cite{kolotouros2019learning, kocabas2020vibe, kolotouros2019convolutional} that estimate SMPL parameters from a single image, we use GCMR~\cite{kolotouros2019convolutional}, the state-of-the-art method, for our experiments.

\subsection{Occupancy and Texture Fields}
The textured surface can be represented with two components: occupancy and texture. 
Inspired by the previous studies~\cite{mescheder2019occupancy, park2019deepsdf}, we define a deep implicit function that predicts an occupancy probability and texture at a given 3D query point as 
\begin{equation}
{G}_{\theta}(p, C(p))=(\alpha_{p},{\mathbf c}_{p}, {\gamma}_{p}), 
\end{equation}
where $\alpha_{p}$ and ${\mathbf c}_{p}$ denote the occupancy probability and texture, respectively, at the corresponding 3D coordinate $p$. 
Output values are predicted from the network $G_\theta$ given a condition variable $C(p)$.
$\gamma$ is used for blending the predicted RGB value with the one sampled from the source image.
The final color is defined as ${\mathbf c}^{\prime}_{p}=\gamma \cdot S(\mathbf{I}, \pi(p))+(1-\gamma) \cdot {\mathbf c}_{p}$,
where $\pi(p)$ denotes the 2D projection coordinate of $p$ on the source image $\mathbf{I}$ and $S(\cdot)$ indicates the sampling function to sample the value at a query location using the bilinear interpolation.
Given occupancy fields that represent the continuous inside/outside probabilities, the surface can be defined as the level set of an occupancy probability by extracting the iso-surface, \ie $\alpha=0.5$. 

\subsubsection{Pixel/Voxel-aligned Representation}
To infer the 3D textured surface from a single image using an implicit function, first, the pixel-aligned feature \cite{saito2019pifu} is utilized as the condition variable.
Furthermore, to strengthen the reconstruction performance under severe occlusions or large pose variations, the voxel-aligned feature is also integrated with the condition variables~\cite{zheng2021pamir}. 
The volumetric feature can be encoded from the predefined parametric human body model, \ie SMPL~\cite{loper2015smpl}, using a 3D CNN network. 
The condition variable is defined as 
\begin{equation}
{C(p)}= (S({\mathbf F}_I, \pi(p)), S({\mathbf F}_V, p))^T,
\end{equation}
where ${\mathbf F}_I= E_I({\mathbf I})$ indicates the 2D image feature map extracted from the 2D CNN encoder $E_I$, ${\mathbf F}_V = E_V({\mathbf V}_O)$ represents the volumetric feature encoded from the 3D CNN network $E_V$, and 
${\mathbf V}_O$ is the occupancy volume generated by voxelizing the SMPL model predicted from the input image.

To train our model, we sample 3D points around the ground-truth mesh. 
Our model learns to predict the occupancy probabilities and texture by minimizing the per-point reconstruction loss with 3D supervision.
The reconstruction loss is defined as
\begin{equation}
\mathcal{L}_{R}=\frac{1}{n_p^o}\sum_{i=1}^{n_p^o}| {\bf \alpha}_{p_i}-{\bf \alpha}^*_{p_i}|^2
~+~ \frac{1}{n_p^c}\sum_{i=1}^{n_p^c}| {\mathbf c}_{p_i}-{\mathbf c}^*_{p_i}|,
\end{equation}
where $n_p^o$ and $n_p^c$ are the number of sampled points for occupancy and color, respectively.
$p_i$ is the i-th sampled 3D point,
and $\alpha^*$ and ${\mathbf c}^*$ are the ground-truth occupancy and RGB value of $p_i$, respectively. 
To train the color prediction, query points are only sampled near the surface, while query points are sampled around the entire 3D space for learning occupancy probabilities. 
Similarly, the color reconstruction loss is applied to the $\gamma$-composited final color, ${\mathbf c}^{\prime}$.

\subsubsection{Volume Rendering Loss} 
However, as shown in \figref{fig:volmuerender}, we discovered that the network trained only with 3D supervision produces the diffused occupancy that results in undesirable artifacts in the projection image and the reconstructed mesh. 
To avoid such issues, we propose combining the 2D volume rendering loss~\cite{mildenhall2020nerf} with the 3D reconstruction loss by interpreting alpha values in the volume rendering process as occupancy probabilities.
We found that the alpha values follow a similar tendency with occupancy probabilities that are close to zero for points outside the surface while approaching 1 for points inside the surface. 
Similar interpretation with respect to alpha values is attempted in a 3D-aware image generation method~\cite{xu2021generative} to define a 3D surface when only 2D supervisions are provided.
Meanwhile, we consider the alpha values as the occupancy probabilities that are trained with 3D supervisions, which helps in additionally utilizing 2D supervisions, \ie projected images from ground-truth meshes. 
Thus, we formulate the occupancy-based volume rendering as
\begin{equation}
\mathbf{c}_{r}=\sum_{j=1}^{N} \alpha_{j} \prod_{k=1}^{j-1}\left(1-\alpha_{k}\right) \mathbf{c}_{j}, 
\end{equation}
where ${\mathbf r}$ is the ray marched from the randomly selected target view and
$N$ and $j$ are the number and the index of sampled points along a ray, respectively.
The volume rendering loss is defined as 
\begin{equation}
\mathcal{L}_{\text{vol}}=\frac{1}{n_r}\sum_{i=1}^{n_r}| {\mathbf c}_{r}-{\mathbf c}^*_{r}|,
\end{equation}
where $n_r$ is the number of rays, and ${\mathbf c}^*$ is the ground truth RGB value in the target-view image.
The same volume rendering loss is applied to the $\gamma$-composited final color value, ${\mathbf c}^{\prime}$.
We adopt the hierarchical sampling strategy proposed in NeRF~\cite{mildenhall2020nerf} that densely samples points based on the normalized weights of colors along the points on the ray.



\subsection{Refinement and Fusion}
%
Since the geometry and the texture of the initially predicted 3D surface are far short of reality from the backside view, we propose the refine-and-fuse approach that 1) {\it refines} the coarse backside-view image by warping the source feature and 2) {\it fuses} the refined image with the input condition to generate the final textured surface.
%
\subsubsection{Refinement Stage}
After the initial stage,
we can render the coarse image ${\mathbf I}^{\prime}_{trg}$ from the backside view and the warping function ${\mathbf f}$ from the source image to the target image.
%
The warping function is computed by volume-rendering the 3D query points $p$ with the value of source grid coordinates $\pi(p)$ using the initially learned occupancy values.
This warping function maps the coordinates of the source view to those of the target view and transfers the semantically similar visual features, \eg the texture of clothes and skin color, from the front to the back. 

\begin{figure}[t]
\captionsetup{width=1\linewidth}
\begin{center}
\begin{tabular}{@{}c}
\includegraphics[width=1.0\linewidth]{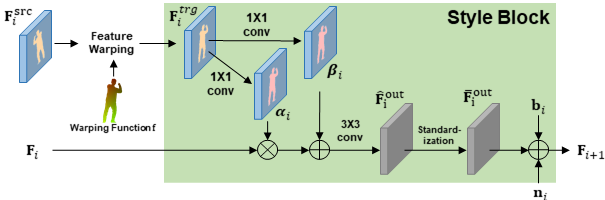} 
\end{tabular}
\end{center}
\caption{\small Style block with feature warping. To spatially modulate the intermediate feature of the i-th style block $\mathbf{F}_i$, we warp the source feature with the warping function ${\mathbf f}$. After being modulated with $\boldsymbol \alpha$ and $\boldsymbol \beta$, the feature is standardized. The bias $\mathbf{b}_i$ and the noise $\mathbf{n}_i$ are then added to the output.
The upsampling layer is added before the 3$\times$3 convolution layer in every two style blocks.
}
\label{fig:styleblock}
\end{figure}

\Figref{fig:framework}(b) shows the overview of our refinement network.
We use the StyleGAN2 architecture~\cite{karras2020analyzing} with the spatial modulation~\cite{park2019semantic, albahar2021pose} that takes the coarse target-view image ${\mathbf I}^{\prime}_{trg}$ as the input and the warped source image feature as the modulation condition.
The source image encoder encodes the source image ${\mathbf I}$ to multi-scale features ${\mathbf F}^{src}_{i}$ with several residual blocks.
Then the source features are warped to the target view 
with the warping function $\mathbf f$ for spatial alignment.
As illustrated in \figref{fig:styleblock}, the warped source feature ${\mathbf F}^{trg}_{i}$ is passed to two 1$\times$1 convolution layers that output affine parameters of the modulation layer, \ie scale ${\boldsymbol \alpha}_i$ and shift ${\boldsymbol \beta}_i$, to perform the spatial modulation.
Unlike the standard modulation of StyleGAN~\cite{karras2019style}, our network utilizes the spatial modulation for the intermediate feature of the i-th style block ${\mathbf F}_i$ to preserve the fine details of a source image.
The modulated feature $\hat{\mathbf F}_i$ is passed to a 3$\times$3 convolution layer of the style block that outputs $\hat{\mathbf F}^{out}_i$ and is then standardized to have zero mean and unit standard deviation. 
The final output of the style block ${\mathbf F}_{i+1}$ is computed by adding the bias and the noise~\cite{karras2020analyzing} to the standardized feature $\bar{\mathbf F}^{out}_i$.
The entire formulation of the style block with feature warping is written as
\begin{equation}
\begin{gathered}
{\mathbf F}^{trg}_{i} = \text{warp}({\mathbf F}^{src}_{i}, {\mathbf f}),\quad  {\boldsymbol \alpha}_{i}, {\boldsymbol \beta}_{i} = E_{\text{affine}}({\mathbf F}^{trg}_{i}),\\
\hat{\mathbf F}_i = {\boldsymbol \alpha}_i \otimes \ {\mathbf F}_{i} + {\boldsymbol \beta}_i, \quad \bar{\mathbf F}^{out}_i = \frac{\hat{\mathbf F}^{out}_i - \mu(\hat{\mathbf F}^{out}_i)}{\sigma(\hat{\mathbf F}^{out}_i)},
{\mathbf F}_{i+1} = \bar{\mathbf F}^{out}_i+{\mathbf b}_i+{\mathbf n}_i 
\end{gathered}
\end{equation}
where $\mu(\hat{\mathbf F}^{out}_i)$ and $\sigma(\hat{\mathbf F}^{out}_i)$ denote the mean and standard deviation of $\hat{\mathbf F}^{out}_i$, and ${\mathbf b}_i$ and ${\mathbf n}_i$ are the bias and the noise~\cite{karras2020analyzing}, respectively.

Our refinement network is trained by optimizing the following objectives: 1) the $\ell_{1}$ loss, 2) the perceptual loss, and 3) the adversarial loss.
We minimize the $\ell_{1}$ loss between the foreground human regions of the refined target-view image $\hat{\mathbf I}_{trg}$ and those of the ground-truth target-view image ${\mathbf I}_{trg}$. 
The reconstruction loss is formulated as
    \begin{equation}
    \mathcal{L}_{\ell_{1}}(\psi)=\left\|\hat{\mathbf I}_{trg} \cdot {\mathbf M}_{trg}-{\mathbf I}_{trg} \cdot {\mathbf M}_{trg}\right\|_{1},
    \end{equation}
where ${\mathbf M}_{trg}$ is the human foreground mask, and $\hat{\mathbf I}_{trg}$ denotes the refined target-view image.
Also, we apply the perceptual loss to our training objective by minimizing the weighted sum of the $\ell_{1}$ loss between the pre-trained VGG features of the refined target-view image $\hat{\mathbf I}_{trg}$ and those of the ground-truth target-view image ${\mathbf I}_{trg}$ as
     \begin{equation}
    \mathcal{L}_{\text{vgg}}(\psi)=\sum_{i=1}^{5} w_{i} \cdot\left\|VGG_{l_{i}}\left(\hat{\mathbf I}_{trg} \cdot {\mathbf M}_{trg}\right)-VGG_{l_{i}}\left({\mathbf I}_{trg} \cdot {\mathbf M}_{trg}\right)\right\|_{1},
    \end{equation}
where $w$ is [$\frac{1}{32}, \frac{1}{16}, \frac{1}{8}, \frac{1}{4}, 1.0$] and $l$ is [1,6,11,20,29]. 

Lastly, we adopt a conditional GAN~\cite{isola2017image} by training a discriminator to match the real and fake images conditioned on the coarse target-view image.
The non-saturating GAN loss~\cite{gan} is used with R1 regularization~\cite{r1_regularization}. 
The entire formulation of the conditional GAN loss is written as
    \begin{equation}
    \begin{split}
        \vspace{-0.3cm}
        \mathcal{L}_\text{adv}(\psi, \phi) = \mathbf{E}[g(D_{\phi}(G_{\psi}({\mathbf I}, {\mathbf I}^{\prime}_{trg}, {\mathbf f}), {\mathbf I}^{\prime}_{trg}))] \\
        + \mathbf{E}[g(-D_{\phi}({\mathbf I}, {\mathbf I}^{\prime}_{trg})) 
        + \lambda \|\nabla D_{\phi}({\mathbf I}, {\mathbf I}^{\prime}_{trg})\|^{2}],
        \vspace{-0.3cm}
    \label{GANloss}
    \end{split}
\end{equation}
where $g(x) = -log(1 + exp(-x))$, $G_{\psi}$ denotes our refinement generator, and $D_{\phi}$ indicates the discriminator. 
$\lambda$ is the hyper-parameter that balances the gradient penalty effect.

Our full training objective functions for the refinement network $G_{\psi}$ are written as 
\begin{equation}
\begin{aligned}
&\mathcal{L}_{\text{total}}=\mathcal{L}_{\text{adv}}+\lambda_{\text{vgg}}\mathcal{L}_{\text{vgg}}+\lambda_{\ell_{1}}\mathcal{L}_{\ell_{1}}, 
\end{aligned}
\end{equation}
where $\lambda_{vgg}$ and $\lambda_{\ell_{1}}$ are hyperparameters determining the importance of each loss.

\subsubsection{Fusion Stage}
To generate the final output utilizing the refined backside-view image, we train the fusion network that takes a pair of front and backside-view images as an input.
Instead of the output of our refinement network, \ie the refined image from the coarse backside-view image, the ground-truth backside-view image is used for training.
Unlike the initial stage, the pixel-aligned feature is additionally extracted from the backside-view image and is then concatenated to the pixel-aligned feature from the source image. 
The condition variable is defined as
\begin{equation}
{C(p)}= (S({\mathbf F}_I, \pi(p)), S({\mathbf F}_{{\mathbf I}_{trg}}, \pi(p)), S({\mathbf F}_V, p))^T. 
\end{equation}
The $\gamma$ values are also estimated to independently sample the color values from the given source image and the backside-view image. 
The final $\gamma$-composited color value is calculated as
\begin{equation}
\begin{aligned}
{\bf c}^{\prime}_p=\gamma_{1} \cdot S(\mathbf{I}, \pi(p)) + \gamma_{2} \cdot S(\mathbf{I}_{trg}, \pi(p)) + \gamma_{3} \cdot {\bf c}_p,
\end{aligned}
\label{eq:composited}
\end{equation}
where $\gamma$ values are normalized with the softmax function.
Except for the input condition and the $\gamma$ compositing, the training process, including the network architecture and the loss function, is identical to the initial stage.

\section{Experiments}


\subsection{Experimental Setup}
\subsubsection{Dataset}
For training and evaluating our method, we collected 526 samples from the THUman2.0~\cite{zheng2021deepmulticap} dataset and 498 samples from the Twindom dataset that include high-quality textured human scans. 
50 samples from the THUman2.0 and Twindom datasets were used for evaluation and the remaining samples were used for training (476 samples from THUman2.0 and 448 samples from Twindom).
1,396 real-world full-body images collected from DeepFashion~\cite{liu2016deepfashion} were also utilized for evaluations. 

To preprocess the training data, we followed PaMIR~\cite{zheng2021pamir} by rendering the training scans from 360 viewpoints at a resolution of 512$\times$512.
MuVS~\cite{huang2017towards} was applied to the rendered multi-view images for preparing the ground-truth SMPL model. 
For evaluation, the evaluation set of 3D scans from 4 views spanning every 90 degrees in the horizontal axis were rendered. 

\subsubsection{Training/Evaluation Setup}
For the initial stage, the network is trained with a learning rate of $2\times10^{-4}$ using the ADAM optimizer with $\beta_{1}=0.9$ and $\beta_{2}=0.999$.
The network is trained for 200,000 iterations with a batch size of 3, and 5,000 points are sampled 
both for occupancy and color reconstruction. 
For optimizing volume rendering loss, the camera locations are randomly sampled among 360 degrees around the object. 
1,000 rays are cast for each sampled camera location and 48 samples are used per ray adopting hierarchical sampling.
Instead of using the predicted SMPL models~\cite{zheng2021pamir}, we use the ground-truth ones during training since we empirically found that using the predicted SMPL model causes unstable training due to large pose variations of THUman2.0.  
The pre-trained GCMR is used and body reference SMPL optimization~\cite{zheng2021pamir} is performed for more accurate SMPL input in the evaluation.
The fusion stage follows the same experimental setup as the initial stage, except for a batch size of 2.

For the refinement network, different hyper-parameter values are set for the generator and the discriminator. 
The network is trained with a batch size of 1, a learning rate of $2\times10^{-3} \times ratio$, and the ADAM optimizer with $\beta_{1}=0$ and $\beta_{2}=0.99^{ratio}$.
The $ratio$ is set to $\frac{4}{5}$ for the generator and $\frac{16}{17}$ for the discriminator.
We train the initial stage first and then train the refinement stage using the output of the initial stage.


\subsection{Experimental Results}
Here, we compare our method with the state-of-the-art methods 
and demonstrate that our method produces more realistic 3D human reconstruction from a single image. 
We select PIFu~\cite{saito2019pifu} and PaMIR~\cite{zheng2021pamir} as our baselines which are single-image 3D human reconstruction methods based on an implicit function.
Note that we omit the comparisons where textures cannot be estimated and open-source codes are not available.
The output meshes are extracted using the Marching Cube algorithm at an occupancy threshold of 0.5
and textured by predicting the colors of the extracted vertices. 
All comparison methods are retrained on our training datasets. 

\subsubsection{Qualitative Comparisons}
\begin{figure*}[t]
\captionsetup{width=1\linewidth}
\centering
\def\arraystretch{0.2}
\begin{tabular}{@{}c}
\includegraphics[width=0.9\linewidth]{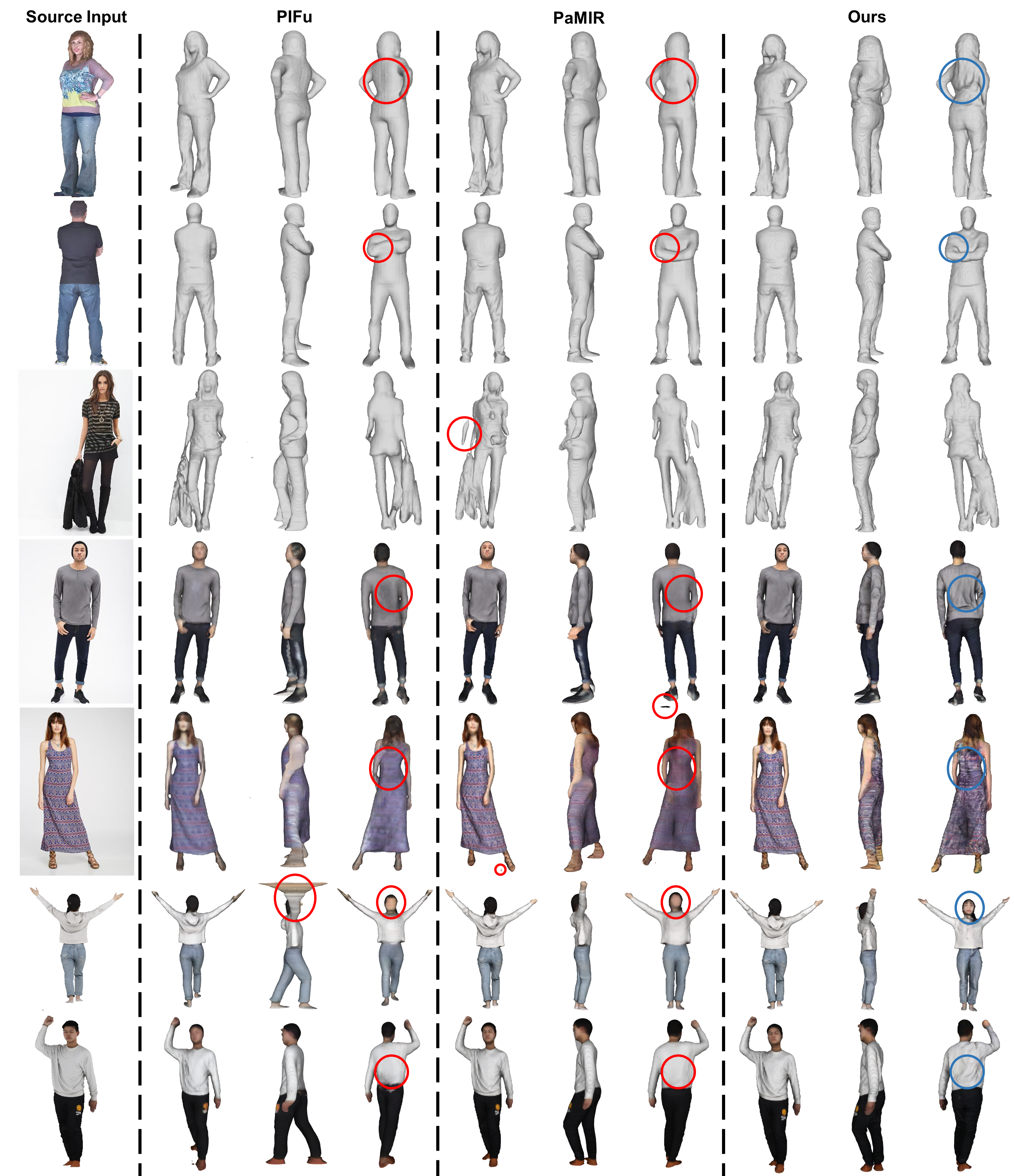} \\
\end{tabular}
\caption{\small Qualitative comparison of output meshes on the THUman2.0, Twindom, and DeepFashion datasets. Given the source image in the first column, meshes without texture are presented in rows 1-3, and textured meshes are presented in rows 4-7. The output meshes are captured from the front, side, and back views for each method.}
\label{fig:quali-condition}
\end{figure*}

Qualitative comparisons are shown in \Figref{fig:quali-condition}. 
As shown in the figure, our method shows the best visual quality from any view in terms of surface geometry and texture. 
PIFu~\cite{saito2019pifu} has difficulty in reconstructing human bodies in challenging poses and suffers from depth ambiguity (see row 6) since it does not utilize any prior knowledge of human shape as a condition. 
%
%
While PaMIR~\cite{zheng2021pamir} succeeds in reconstructing a full-body model under challenging poses, it still suffers from floating artifacts (row 3).
Both compared methods represent an over-smoothed textured surface, especially in the unseen regions, for all examples. 
However, as noticeable in the first row of \Figref{fig:quali-condition}, our method shows the backside view with high-frequency details such as wrinkles in clothes and hair.
Our method is also capable of preserving patterns of clothes by warping the source features (row 5) and generating the face when the backside of the human body is captured for the source input image (row 6).
Thanks to our refine-and-fuse approach, our method exhibits the realistic quality of both surface geometry and texture from the unobserved view. 
%


\subsubsection{Quantitative Comparisons}
We quantitatively evaluated our method by measuring the geometry reconstruction accuracy and texture quality.
To measure the geometry reconstruction, we followed the previous studies~\cite{saito2019pifu, zheng2021pamir}, by measuring the point-to-surface (P2S) error and the Chamfer distance between the predicted mesh and the ground-truth one.
The quantitative results are presented in \tabref{tb:quanti}.
%
%
%
Although our method shows outstanding qualitative results
(see \figref{fig:quali-condition}), it shows marginal improvement compared to the baselines because these distance-based metrics cannot measure the perceptual quality.

\begin{table}[t]
\caption{\small Quantitative comparison of surface geometry reconstruction on the THUman2.0 and Twindom datasets.}
\centering
\renewcommand{\tabcolsep}{2mm}
\renewcommand{\arraystretch}{1.1}
\captionsetup{width=1\linewidth}
{\scriptsize
%
\begin{tabular}{c|c|c|c|c}

\hline
 & \multicolumn{2}{c|}{THUman2.0} & \multicolumn{2}{c}{Twindom} \\ \hline
Method & P2S↓ (cm) & Chamfer↓ (cm) & P2S↓ (cm)& Chamfer↓ (cm) \\ \hline
PIFu~\cite{saito2019pifu} &1.997  & 1.807  & 1.199  & 1.198 \\ \hline
PaMIR~\cite{zheng2021pamir}  & 1.346  &  \bf 1.417  & 0.949  &  0.950  \\ \hline
Ours (Initial)  & 1.358   &  1.459   & 0.941  & 0.953  \\ \hline
Ours &\bf 1.339 & 1.440 & \bf 0.923 &  \bf 0.940  \\ \hline
\end{tabular}%
}
\label{tb:quanti}
\end{table}

To further evaluate the texture quality of our approach, we measured the peak signal-to-noise ratio (PSNR)
and the learned perceptual image patch similarity (LPIPS). 
PSNR measures the distortion and LPIPS measures the perceptual similarity between images. 
We also introduce the ReID distance to evaluate the perceptual similarity between the source image and the predicted backside-view image. 
Using the pre-trained encoder for the person re-identification task~\cite{zhou2019omni}, the cosine distance is computed between the features extracted from the source image and the generated backside-view image. 
To evaluate the pure texture quality without geometric errors, the textured mesh is generated with the ground-truth vertices by predicting their vertex colors.
Note that the ground-truth SMPL is used for the input of both PaMIR and our method.
We then render the backside color image of the output textured mesh by projecting to the reversed viewpoints of the source views without a light condition.
The quantitative results for texture are shown in \tabref{tb:quanti_texture}. 
As shown in the table, \ourmodel~achieves the best LPIPS and ReID distance by a large margin, which implies that our method predicts realistic and source-consistent texture. 

\begin{table*}[t]
\caption{\small Quantitative comparison of texture quality on the THUman2.0, Twindom, and DeepFashion datasets. The texture quality is presented by rendering the predicted vertex colors of the ground-truth mesh vertices. Since the  ground-truth mesh vertices of the DeepFashion dataset do not exist, the colors of the predicted mesh vertices are rendered.}
\centering
\renewcommand{\tabcolsep}{3mm}
\renewcommand{\arraystretch}{1.1}
\captionsetup{width=1\linewidth}
{\scriptsize
%
\begin{tabular}{c|c|c|c|c|c|c|c}
\hline
 & \multicolumn{3}{c|}{THUman2.0} & \multicolumn{3}{c|}{Twindom} &\multicolumn{1}{c}{DeepFashion} \\ \hline
Method & PSNR↑ & LPIPS↓ & ReID dist↓ & PSNR↑ & LPIPS↓ & ReID dist↓ & ReID dist↓  \\ \hline
PIFu~\cite{saito2019pifu} & 25.13  & 0.095  & 0.199 &24.60  & 0.089  & 0.194 & 0.269  \\ \hline
PaMIR~\cite{zheng2021pamir} & \bf 26.62  & 0.078  & 0.163 &24.18  & 0.081  & 0.176 &  0.271  \\ \hline
Ours (Initial) & 26.46  & 0.079  & 0.167 & 24.68  & 0.079  & 0.171  &  0.271 \\ \hline
Ours & 25.97 & \bf 0.074 &\bf 0.139 & \bf25.77  & \bf 0.077  & \bf 0.148 & \bf 0.227   \\ \hline
\end{tabular}%

}
\label{tb:quanti_texture}
\end{table*}







\subsection{Ablation Study}

\subsubsection{Refine-and-Fuse Approach}
\label{sec:ablation_stage}
Here, we quantitatively and qualitatively demonstrate the effectiveness of our refine-and-fuse methodology.
The quantitative comparison between the initial and the final outputs in terms of surface geometry and texture are shown in \tabref{tb:quanti} and \tabref{tb:quanti_texture}, respectively.
Our refine-and-fuse approach improves the results of the initial stage in P2S and the Chamfer distance in \tabref{tb:quanti}, which indicates our methodology enhances the geometry reconstruction quality by alleviating the ambiguity of the backside shape.
Also, our final outputs present a perceptually improved texture quality showing a lower LPIPS and ReID distance than the ones of the initial stage in \tabref{tb:quanti_texture}.

\begin{figure}[h]
\captionsetup{width=1\linewidth}
\centering
\def\arraystretch{0.2}
\begin{tabular}{@{}c}
\includegraphics[width=0.85\linewidth]{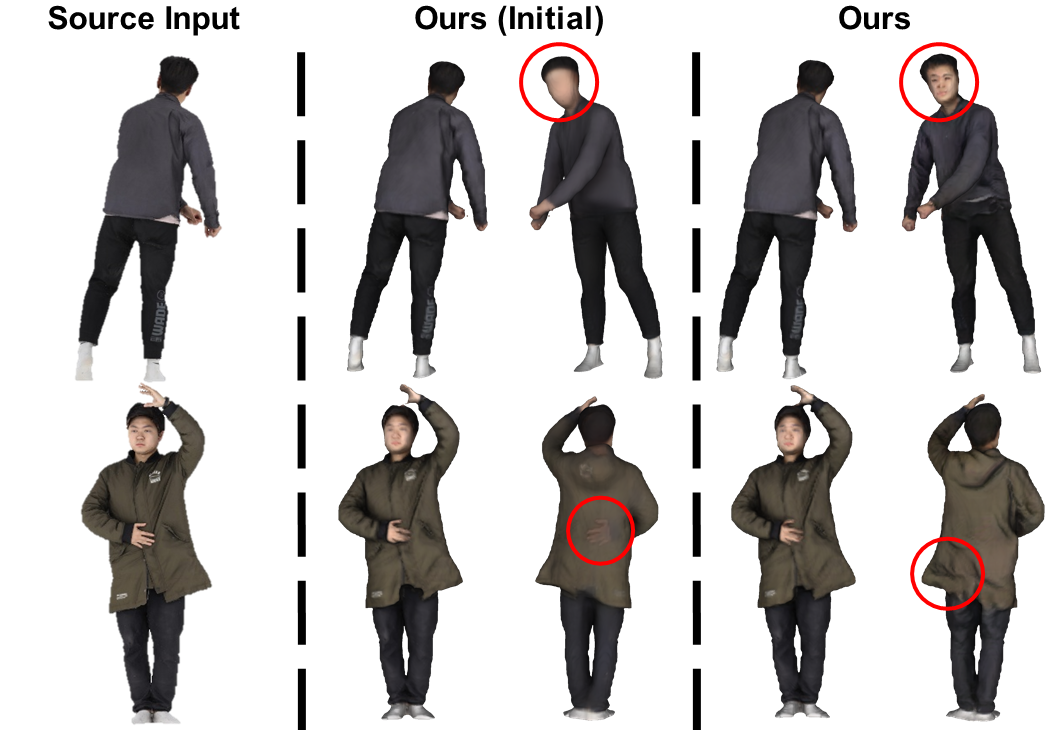} \\
\end{tabular}
\caption{\small Ablation study of our refine-and-fuse approach. The images of textured meshes captured from the front and backside views are presented.}
\label{fig:quali-ablation}
\end{figure}

The qualitative differences between the outputs of the initial and the fusion stage are presented in \Figref{fig:quali-ablation}. 
Our fusion network successfully generates realistic details such as faces and cloth wrinkles, while the initial outputs show an over-smoothed surface on the backside.
Also, the undesired artifacts copied from the source image are removed after refinement (row 2). 


\subsubsection{Spatial Modulation with Source Feature Warping}
\label{sec:stage2_ablation}
%
%
We analyze the effectiveness of the source feature and the spatial modulation with the front-to-back warping function in the refinement stage.
Our refinement network is compared with two variants: (a) spatial modulation with the coarse target feature and (b) AdaIN~\cite{huang2017arbitrary} modulation with the source feature. 
For variant (b), the condition of AdaIN is defined as the style vector that is the average-pooled output of the source image encoder.
As shown in \Figref{fig:quali-condition-ablation},
%
variant (a) fails to generate realistic texture since the high-frequency information is not provided from any real images.
Variant (b) cannot preserve the local texture details in the source image, such as patterns in clothes, as the condition of modulation loses the spatial information of the source feature.
Thanks to our spatial modulation with the source feature warping, our full refinement network effectively generates realistic images by preserving the local details in the source image.

\begin{table}[t]
\caption{\small Ablation study of the source feature and the spatial modulation with feature warping. Note that the texture quality on the output images of the refinement stage are measured.}
\centering
\renewcommand{\tabcolsep}{1.5mm}
\renewcommand{\arraystretch}{1.1}
{\scriptsize
\begin{tabular}{c|c|c|c|c|c|c|c}
\hline
\multicolumn{1}{c|}{} & \multicolumn{3}{c|}{THUman2.0} & \multicolumn{3}{c|}{Twindom} &\multicolumn{1}{c}{DeepFashion} \\ \hline
Method & PSNR↑ & LPIPS↓ & ReID dist↓ & PSNR↑ & LPIPS↓ & ReID dist↓ & ReID dist↓  \\ \hline
 (a) & 25.23  & 0.065  & 0.168 &24.78  &0.065 &0.157& 0.208  \\ \hline
 (b) & 25.75 & 0.059  & 0.158  &25.48 &0.058 &0.148& 0.199   \\ \hline
 Ours & \bf 26.06 & \bf 0.053 &\bf0.143  &\bf 25.85 &\bf 0.052 &\bf 0.126 & \bf 0.167 \\ \hline
\end{tabular}%
}
\label{tb:quanti_ablation}
\end{table}



\begin{figure}[h]
\captionsetup{width=1\linewidth}
\centering
\def\arraystretch{0.2}
\begin{tabular}{@{}c}
\includegraphics[width=0.85\linewidth]{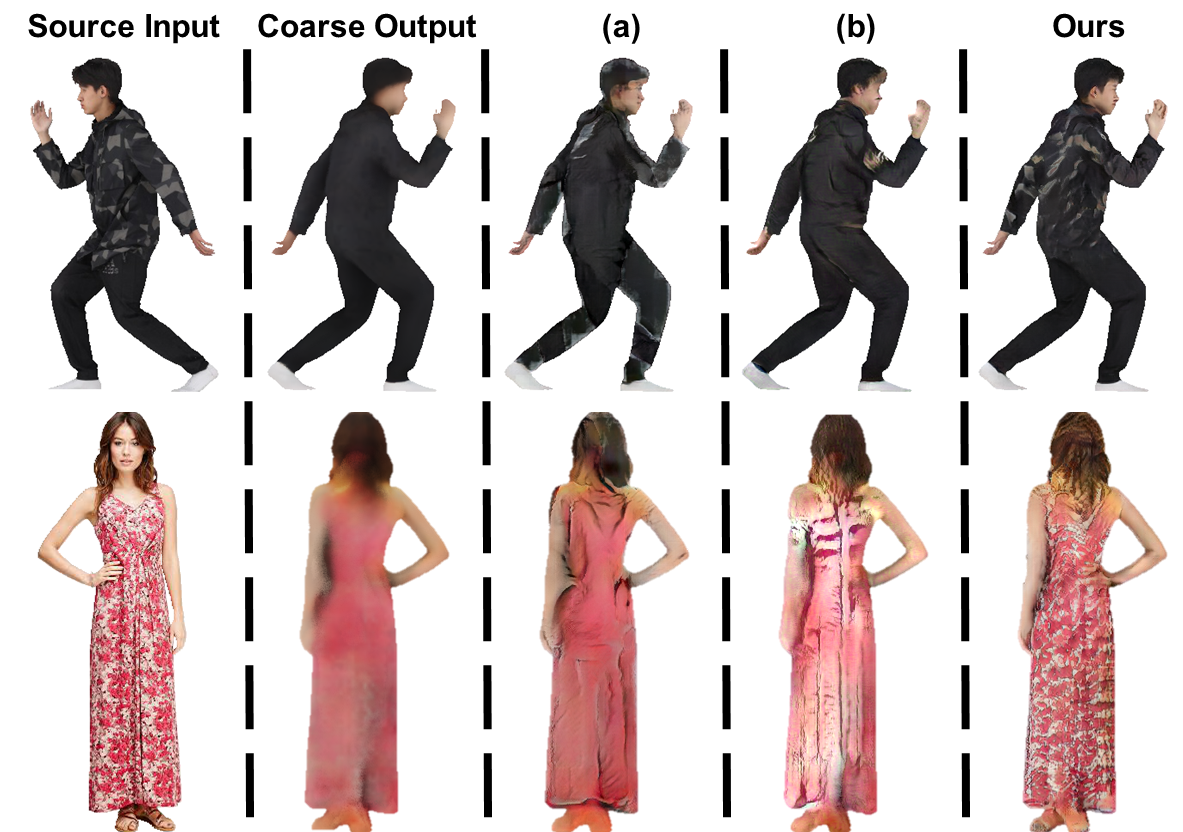} \\
\end{tabular}
\caption{\small Ablation study on the source feature and spatial modulation with warping in the refinement network. Given the source image and the coarse target-view image in columns 1-2, we compare the refined images with those of the variants. Detailed experimental settings for the variants are found in \secref{sec:stage2_ablation}.}
\label{fig:quali-condition-ablation}
\end{figure}



\section{Limitations and Conclusions}
\label{sec:conclusion}

Although the parametric body model provides strong prior knowledge of a human body, a performance bottleneck may occur based on the predicted SMPL accuracy. 
The large errors of the SMPL prediction on challenging poses often lead to failure cases, the examples of which are presented in the supplementary material.
Improving the performance of the SMPL prediction model may boost the output quality of our model. 
Also, by utilizing multi-modal data such as silhouettes, segmentations, and 2D/3D keypoints from pre-trained networks, we expect more accurate inference-stage SMPL optimization compared to body reference optimization~\cite{zheng2021pamir}.

Our model achieves successful results; however, its training and inference process is complicated due to the staged methodology.
In another trial, we attempted single-stage learning where the volume-rendered image is refined while training the initial occupancy and texture. 
However, the volume rendering process of a high-resolution image disrupts the training process, consuming substantial computational time and memory.
We plan to address this issue in our future work by considering more memory-efficient 3D representations or volume-rendering techniques.


In this study, we propose \ourmodel~that {\it refines} the projected backside-view image and {\it fuses} the refined image for predicting the final human body in a coarse-to-fine manner to improve over-smoothed surface geometry and blurry texture from an unobserved view.
To alleviate the artifacts in the projected images and the reconstructed meshes, we propose training the occupancy probability by concurrently leveraging 2D and 3D supervisions with the occupancy-based volume rendering.
In addition, we designed a refinement network that generates backside-view images with fine details using the front-to-back warping function. 
The proposed method achieves state-of-the-art performance in 3D human reconstruction from a single image in terms of surface geometry and texture quality. 
\begin{acks}
This work was supported by FNS HOLDINGS and the Institute of Information \& communications Technology Planning \& Evaluation (IITP) grant funded by the Korean government(MSIT) (No. 2019-0-00075, Artificial Intelligence Graduate School Program(KAIST), No. 2021-0-01778, Development of human image synthesis and discrimination technology below the perceptual threshold, No.2021-0-02068, Artificial Intelligence Innovation Hub).

\end{acks}

\bibliographystyle{ACM-Reference-Format}
\bibliography{reference}

\appendix








\clearpage
\setcounter{page}{1}

\renewcommand{\thefigure}{\Alph{figure}}
\renewcommand{\thetable}{\Alph{table}}



\section{Supplementary: Overview}

Here, we present implementation details and additional experiments which support the validity of our proposed method. We provide details of our network architectures in \secref{sec:network_detail}. We also show additional experiments including the $\gamma$ visualization and target view expansion in \secref{sec:additionalexp}. 
Lastly, we present the failure cases and the additional results of our method in \secref{sec:additionalout}. 

\section{Network Architecture Detail}

\label{sec:network_detail}
\subsection{Occupancy and Texture Network}

The occupancy and texture network,\ie the networks of the initial stage and the fusion stage, consist of feature extractors and a Multi-Layer Perceptrons (MLP) network. 
For the image feature extractor, we follow PIFu~\cite{saito2019pifu} by adapting the stacked hourglass network~\cite{newell2016stacked} that outputs a 256-channel feature map with a resolution of 128$\times$128. 
We also adopt the same 3D volume encoder in PaMIR~\cite{zheng2021pamir} that outputs a 32-channel volume feature with a resolution of 32$\times$32$\times$32.
As shown in Fig. 3 in the main paper, after extracting features from the input image and the SMPL model, the MLP-based network predicts the occupancy probability, color, and $\gamma$ conditioned on these extracted features. 
We concatenate the query coordinate to the input features after applying the positional encoding~\cite{mildenhall2020nerf}. 
The numbers of intermediate channels are (1024,512,256,128), and the input feature is concatenated to each intermediate channel. 
The final output has 5 channels for the network in the initial stage and 7 channels for the fusion network to predict the RGB, occupancy, and $\gamma$ values.

\subsection{Refinement Network}
\label{sec:refinearch}
\begin{figure}[h]
\captionsetup{width=1\linewidth}
\begin{center}
\begin{tabular}{@{}c}
\includegraphics[width=1.0\linewidth]{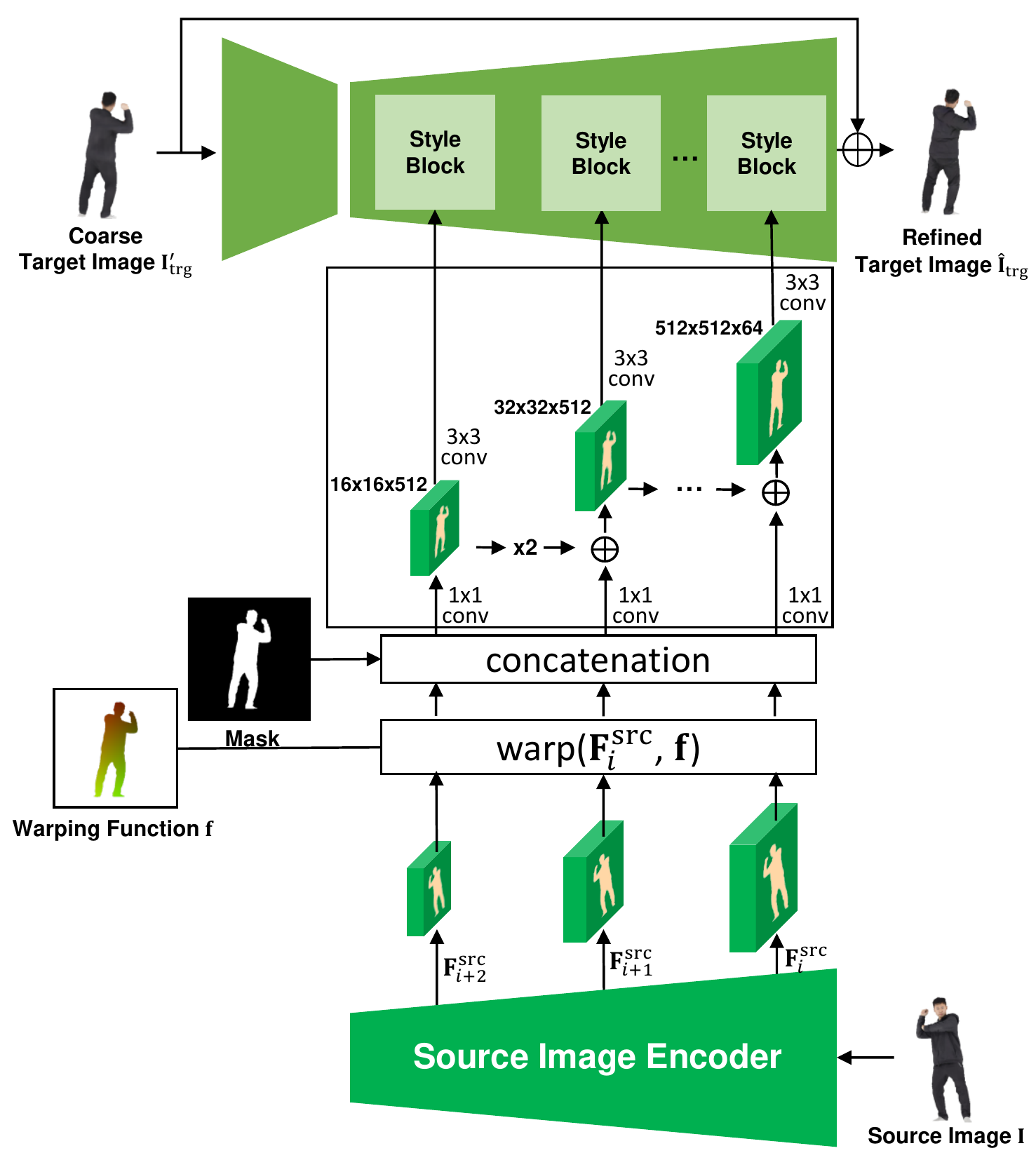} 
\end{tabular}
\end{center}
\caption{\small The detailed network architecture of our refinement network. The multi-scale outputs of the source image encoder are warped to the target image space with the warping function. Utilizing a feature pyramid network~\cite{lin2017feature}, the warped source features are fed into the style blocks as the spatial-modulation conditions. 
}

\label{fig:msfeature}
\end{figure}

As illustrated in \figref{fig:msfeature}, we present the fine details of the refinement network.
We utilize the StyleGAN2 architecture~\cite{karras2019style} with a spatial modulation~\cite{park2019semantic, albahar2021pose} that takes the coarse target-view image ${\mathbf I}^{\prime}_{trg}$ as an input and the warped source image feature as a modulation condition.
The source image ${\mathbf I}$ is encoded to multi-scale features ${\mathbf F}^{src}_{i}$ by the source image encoder that consists of several residual blocks.
For simplification, the main paper explains that the warped source feature ${\mathbf F}^{trg}_{i}$, \ie the result of warping the source feature ${\mathbf F}^{src}_{i}$ with the warping function $\mathbf f$, is passed to the style block. 
In the implementation, the multi-scale warped source features ${\mathbf F}^{trg}_{i}$ are processed with a feature pyramid network~\cite{lin2017feature}. As illustrated in \figref{fig:msfeature}, the warped source features are concatenated to the resized foreground mask, and then passed to several convolution and up-sampling layers.
The outputs of the feature pyramid network are utilized as the modulation conditions of the style blocks.
The ground-truth foreground mask of a target image is used in the training, while the mask is rendered in the inference stage by volume-rendering the occupancy values from the target view and thresholding them with the value of 0.5.

We designed a discriminator that distinguishes real and fake images conditioned on the coarse target-view image by adopting the architecture of StyleGAN2~\cite{karras2020analyzing}.
The architecture details are presented in \tabref{tb:discriminator}.




\begin{table}[h!]
\caption{The configuration of discriminator $D_\phi$.}
\centering
\resizebox{0.85\linewidth}{!}
{
\begin{tabular}{cccc}
\hline
Type & Activation & Input Shape & Output Shape \\ \hline\hline
Input Image and Condition & - & 6$\times$512$\times$512 & 6$\times$512$\times$512 \\ \hline
Conv (1$\times$1) & LeakyReLU (0.2) & 6$\times$512$\times$512 & 64$\times$512$\times$512 \\
\hline
Conv (3$\times$3) & LeakyReLU (0.2) & 64$\times$512$\times$512 & 64$\times$512$\times$512 \\
Conv (3$\times$3, Stride 2) & LeakyReLU (0.2) & 64$\times$512$\times$512 & 128$\times$256$\times$256 \\
Residual Connection & - & 128$\times$256$\times$256 & 128$\times$256$\times$256 \\ 
 \hline
Conv (3$\times$3) & LeakyReLU (0.2) & 128$\times$256$\times$256 & 128$\times$256$\times$256 \\
Conv (3$\times$3, Stride 2) & LeakyReLU (0.2) & 128$\times$256$\times$256 & 256$\times$128$\times$128 \\
Residual Connection & - & 256$\times$128$\times$128 & 256$\times$128$\times$128 \\ 
 \hline
 Conv (3$\times$3) & LeakyReLU (0.2) & 256$\times$128$\times$128 & 256$\times$128$\times$128 \\
 Conv (3$\times$3, Stride 2) & LeakyReLU (0.2) & 256$\times$128$\times$128 & 512$\times$64$\times$64 \\
 Residual Connection & - & 512$\times$64$\times$64 & 512$\times$64$\times$64 \\ 
 \hline
 Conv (3$\times$3) & LeakyReLU (0.2) & 512$\times$64$\times$64 & 512$\times$64$\times$64 \\
 Conv (3$\times$3, Stride 2) & LeakyReLU (0.2) & 512$\times$64$\times$64 & 512$\times$32$\times$32 \\
 Residual Connection & - & 512$\times$32$\times$32 & 512$\times$32$\times$32 \\ 
 \hline
 Conv (3$\times$3) & LeakyReLU (0.2) & 512$\times$32$\times$32& 512$\times$32$\times$32 \\
 Conv (3$\times$3, Stride 2) & LeakyReLU (0.2) & 512$\times$32$\times$32 & 512$\times$16$\times$16 \\
 Residual Connection & - & 512$\times$16$\times$16 & 512$\times$16$\times$16 \\ 
 \hline
 Conv (3$\times$3) & LeakyReLU (0.2) & 512$\times$16$\times$16& 512$\times$16$\times$16 \\
 Conv (3$\times$3, Stride 2) & LeakyReLU (0.2) & 512$\times$16$\times$16 & 512$\times$8$\times$8 \\
 Residual Connection & - & 512$\times$8$\times$8 & 512$\times$8$\times$8 \\ 
 \hline
 Conv (3$\times$3) & LeakyReLU (0.2) & 512$\times$8$\times$8& 512$\times$8$\times$8 \\
 Conv (3$\times$3, Stride 2) & LeakyReLU (0.2) & 512$\times$8$\times$8 & 512$\times$4$\times$4 \\
 Residual Connection & - & 512$\times$4$\times$4 & 512$\times$4$\times$4 \\ 
 \hline
Minibatch stddev & - & 512$\times$4$\times$4 & 513$\times$4$\times$4 \\ \hline
Conv2d (3$\times$3) & LeakyReLU (0.2) & 513$\times$4$\times$4 & 512$\times$4$\times$4 \\ \hline
Conv2d (3$\times$3) & LeakyReLU (0.2) & 513$\times$4$\times$4 & 512$\times$4$\times$4 \\ \hline
Reshape & - & 512$\times$4$\times$4 & 8192 \\ \hline
Linear & LeakyReLU (0.2) & 8192 & 512 \\ \hline
Linear & - & 512 & 1 \\ \hline
\end{tabular}
}
\label{tb:discriminator}
\end{table}

\begin{figure*}[t]
\captionsetup{width=1\linewidth}
\begin{center}
\begin{tabular}{@{}c}
\includegraphics[width=0.7\linewidth]{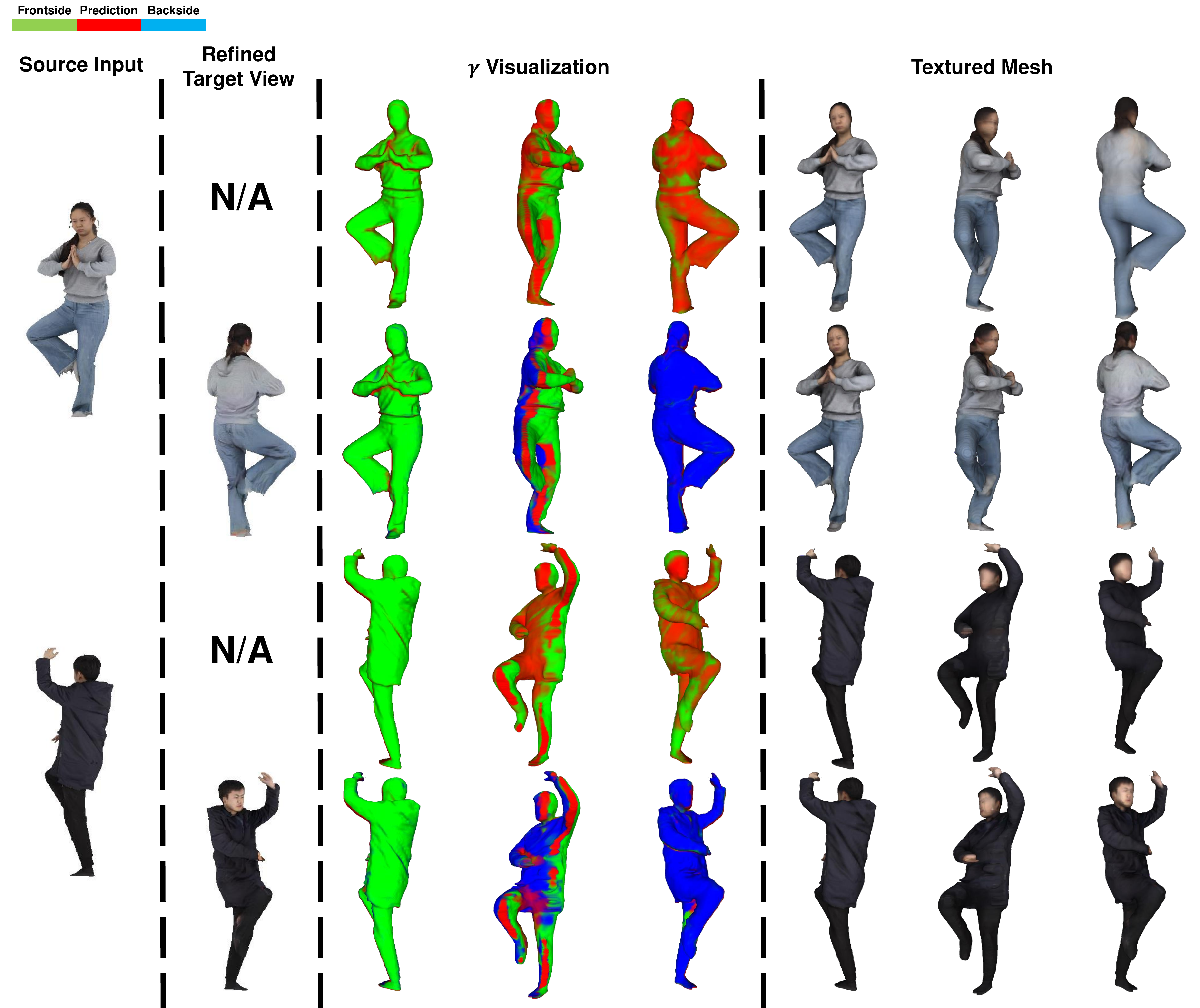} 
\end{tabular}
\end{center}
\caption{\small The $\gamma$ visualization. Given the source input image and the refined target-view image in the first and second columns, the $\gamma$ values predicted from the networks are visualized. 
For each example,the results of the initial stage are presented in the first row and the results of the fusion stage in the second row. 
We visualize the values of $\gamma$ for the prediction, frontside, and backside with red, green, and blue colors, respectively. 
}
\label{fig:gamma}
\end{figure*}

\section{Additional Experiments}
\label{sec:additionalexp}


In this section, we show additional experiments including the $\gamma$ visualization and the target view expansion.

\subsection{$\gamma$ Visualization}
\label{sec:gamma}
We visualize the $\gamma$ value that is used to blend the predicted color value with the ones sampled from the source image and the backside-view image. As shown in \figref{fig:gamma}, given a source image in the first column, the generated backside-view image is shown in the second column, and the meshes captured from the front, side, and back views are presented in the following columns.
For each example, the results of the initial stage are presented in the first row and the results of the fusion stage in the second row. 
As described in the main paper,
the final $\gamma$-composited color value is calculated as Eq. (13).
We visualize the $\gamma_{1}$ in green, $\gamma_{2}$ in blue, and $\gamma_{3}$ in red. 
Note that $\gamma_{2}$ is 0 and $\gamma_{3}$ is $1-\gamma_{1}$ in the initial stage.
We confirm that the $\gamma$ is appropriately predicted, as the $\gamma_{1}$ and $\gamma_{2}$ are close to 1 at the locations visible from the source view and target view, respectively. 

\subsection{Target View Expansion}
\label{sec:multiview}
\begin{figure*}[h]
\captionsetup{width=1.0\linewidth}
\begin{center}
\begin{tabular}{@{}c}
\includegraphics[width=0.6\linewidth]{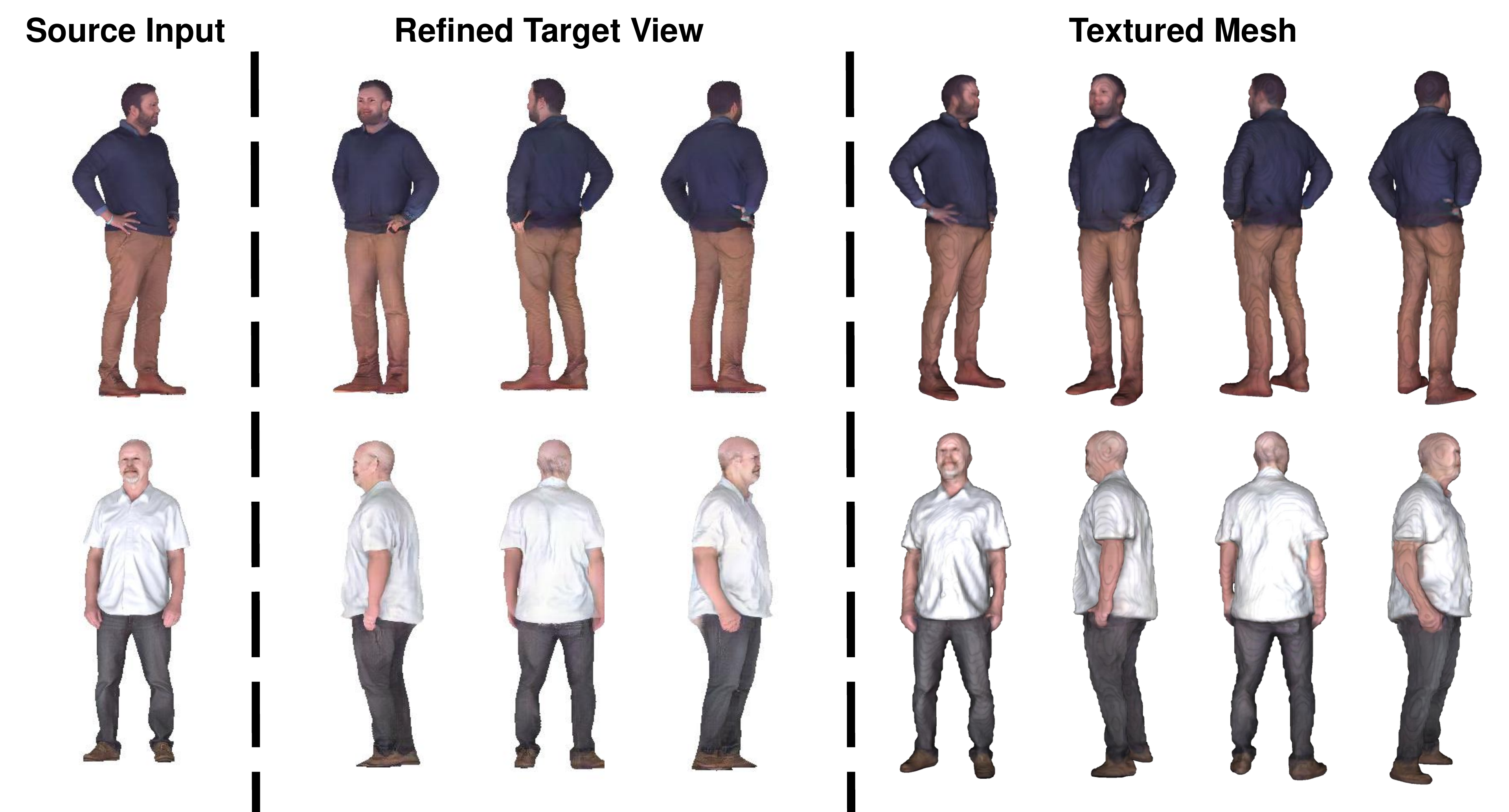} 
\end{tabular}
\end{center}
\caption{\small Qualitative results of the 4-view variant. Given a source image in the first column, refined images from the right, back, and left views are presented in the next column, and the textured meshes captured from the front, side, and back view are presented. 
}
\label{fig:4view}
\end{figure*}

The proposed model is conditioned on two views including the source and the refined backside-view image.
In this section, we demonstrate that our method is expandable to a variant with more target views.
We conducted the experiment of a 4-view variant, \ie the model that utilizes a source-view image and three additional target-view images including the back-, left-, and right-side views. 
An identical refine-and-fuse approach was applied to these additional target-view images.
As shown in \figref{fig:4view}, we report the qualitative results of our 4-view variant.
Compared to the original version that generates the backside-view image only, the 4-view variant directly alleviates the ambiguity of the side view by additionally generating side-view images. 
However, utilizing more than four views harms the computation efficiency without a significant improvement in the perceptual quality. 
This is because the total area of observable region barely increases when using more than 4 views. 
We empirically found that using 2-4 views improves the perceptual quality without degrading the computation efficiency; thus, the 2-view version, \ie the source image with the refined backside-view image, is utilized as our default setting.
%




\begin{figure*}[h]
\captionsetup{width=1.0\linewidth}
\begin{center}
\begin{tabular}{@{}c}
\includegraphics[width=0.7\linewidth]{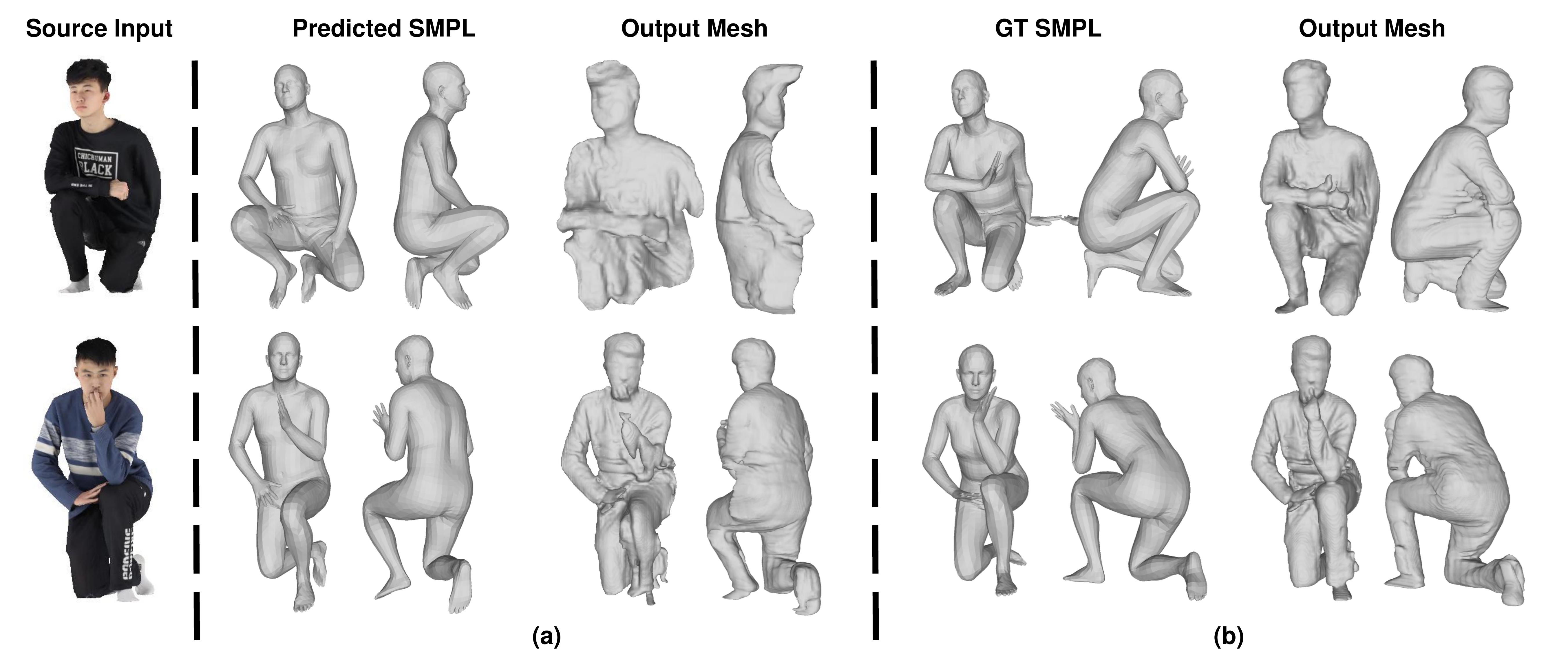} 
\end{tabular}
\end{center}
\caption{\small Failure cases. We present the reconstructed meshes from our method on challenging poses. (a) denotes the outputs when utilizing the predicted SMPL, and (b) denotes those with the ground-truth SMPL.
}
\label{fig:failure}
\end{figure*}



\section{Additional Results}
\label{sec:additionalout}

\subsection{Failure cases}
\label{sec:failure}

This section presents the failure cases of our proposed method.
As described in Sec. 5, despite the fact that the parametric body model provides strong prior knowledge of the human body, a performance bottleneck occurs based on the accuracy of the predicted SMPL.
In practice, SMPL prediction methods~\cite{kolotouros2019convolutional, kolotouros2019learning} show a large error of accuracy for cases of extreme poses, such as sitting, since such poses are rarely seen in their training dataset. 
The imprecise SMPL prediction inevitably degrades the performance of our method resulting in failure cases, as shown in \figref{fig:failure}.
Our method fails to reconstruct accurate meshes for sitting poses since the input SMPL is not aligned to the source image(\figref{fig:failure}(a)).
However, our method shows an improved quality of the mesh when the accurate SMPL input is used (\figref{fig:failure}(b)).

\subsection{Additional Examples}
\label{sec:addexamples}

Here, we include additional visual results to show the mesh quality of our method. 
Additional mesh results with and without texture on the THUman2.0, Twindom, and DeepFashion datasets are presented in  \figref{fig:quali-add}.
The meshes are captured from the front, side, and back views. Given the source image in the first column, our method successfully reconstructs the textured meshes. Note that our method generates high-frequency details such as hair, cloth wrinkles, and faces, even from an unobserved view. Also, the patterns of clothes are generated in a source-consistent manner due to our spatial modulation with feature warping.


\begin{figure*}[t]
\centering
\def\arraystretch{0.2}
\begin{tabular}{@{}c}
\includegraphics[width=0.8\linewidth]{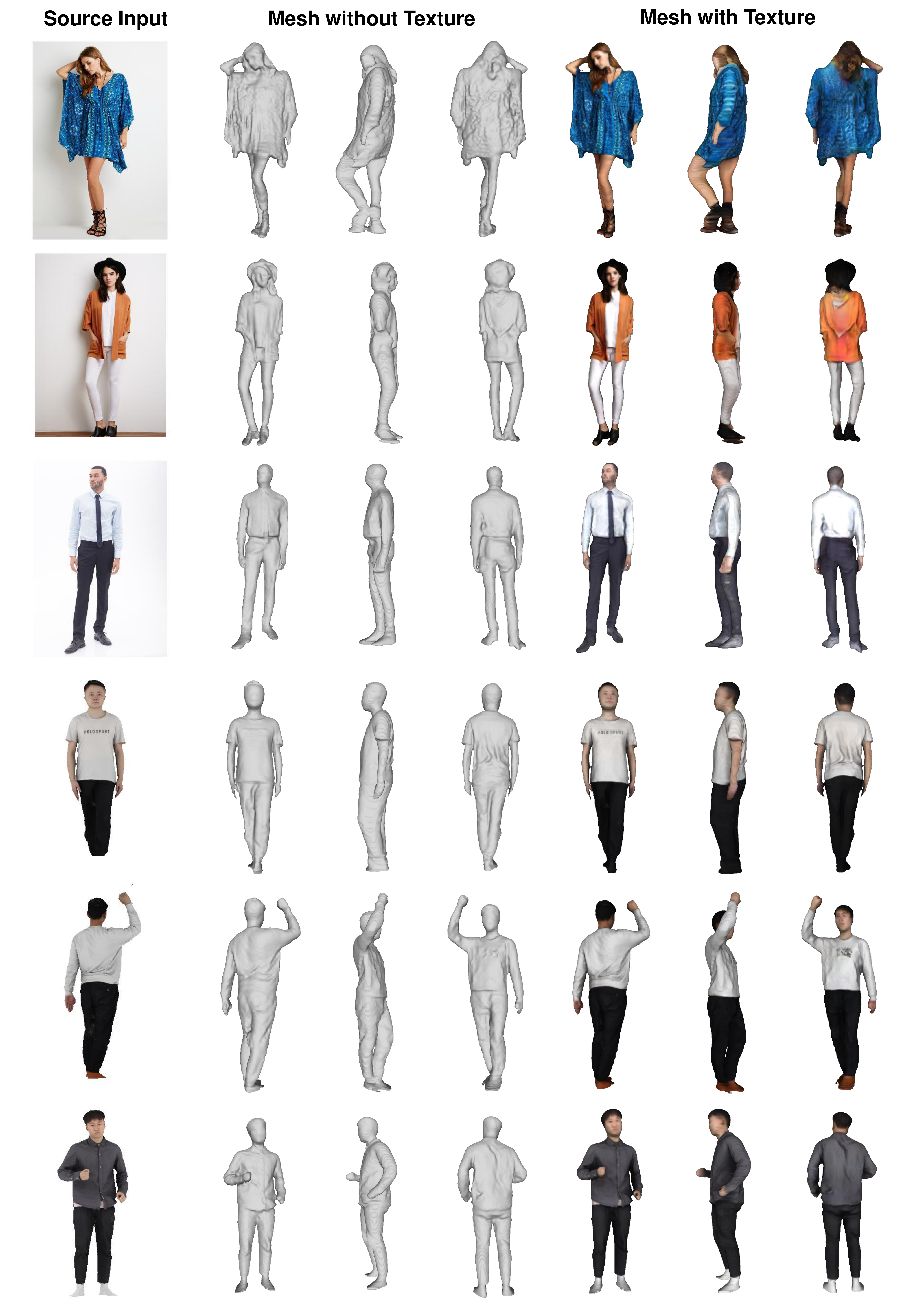} \\
\end{tabular}
\caption{\small Qualitative results of our proposed method. Given the source image in the first column, meshes without texture are presented in columns 1-3, and meshes with texture are presented in columns 4-6. We capture the output meshes from the front, side, and back views (continued.).}
\label{fig:quali-add}
\end{figure*}

\begin{figure*}[t]
\centering
\ContinuedFloat
\def\arraystretch{0.2}
\begin{tabular}{@{}c}
\includegraphics[width=0.8\linewidth]{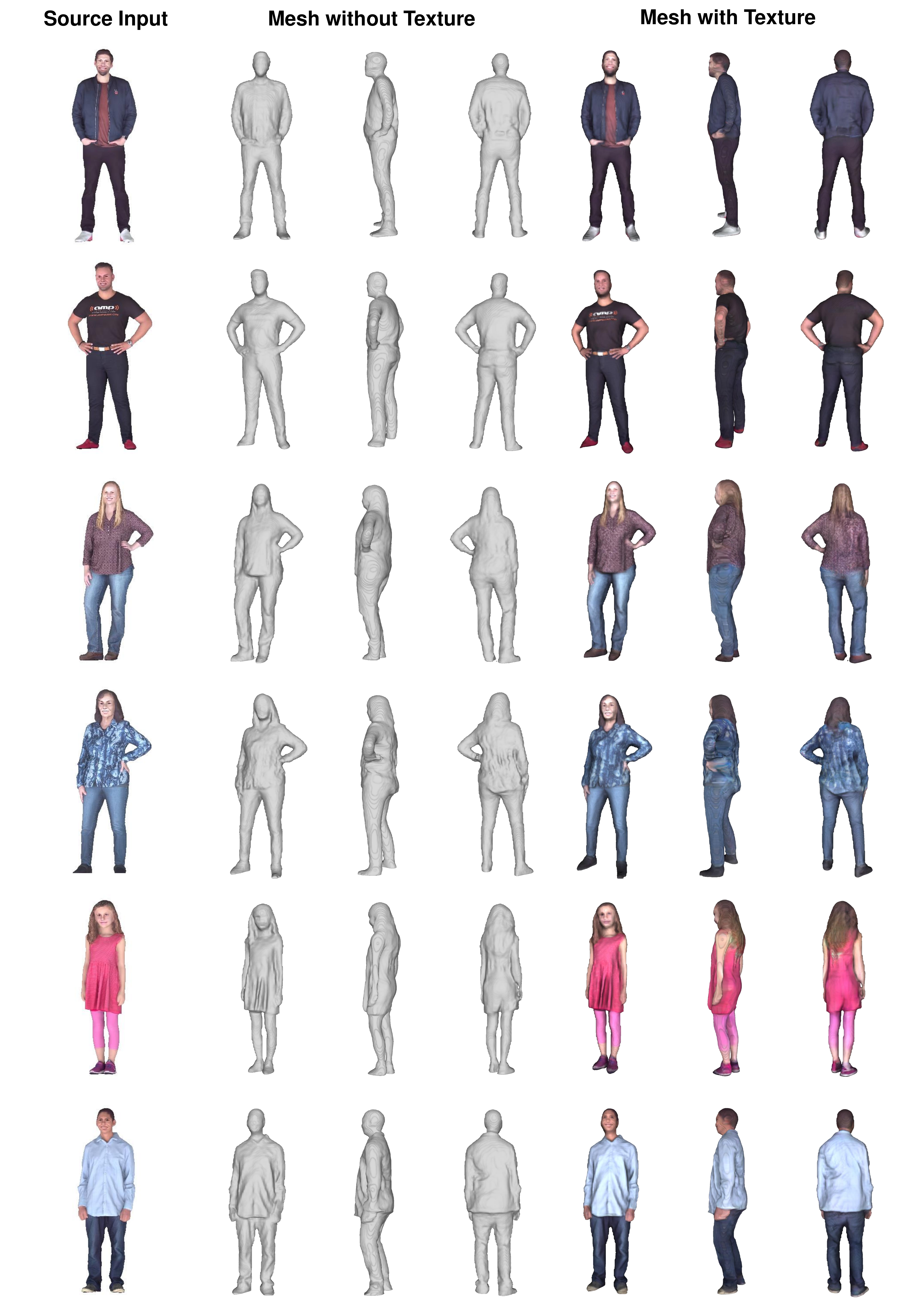} \\
\end{tabular}
\caption{\small  Qualitative results of our proposed method. Given the source image in the first column, meshes without texture are presented in columns 1-3, and meshes with texture are presented in columns 4-6. We capture the output meshes from the front, side, and back views.}
\label{fig:quali-add}
\end{figure*}

\newpage

\clearpage









\end{document}